\pdfoutput=1
\documentclass{article}

\PassOptionsToPackage{numbers, compress}{natbib}

 \usepackage[main, final]{neurips_2026}

\usepackage[utf8]{inputenc} %
\usepackage[T1]{fontenc}    %
\usepackage{hyperref}       %
\usepackage{url}            %
\usepackage{booktabs}       %
\usepackage{multirow}      %
\usepackage{float}         %
\usepackage{wrapfig}       %
\usepackage{graphicx}      %
\usepackage{amsmath}       %
\usepackage{amsfonts}       %
\usepackage{amssymb}        %
\usepackage{mathtools}     %
\usepackage{algorithm}
\usepackage{algpseudocode}
\usepackage{nicefrac}       %
\usepackage{microtype}      %
\usepackage{xcolor}         %
\usepackage{colortbl}       %
\usepackage{xspace}
\usepackage[most]{tcolorbox}
\usepackage{fontawesome5}
\newcommand{\modelname}{\textsc{Vigor}\xspace}

\definecolor{vigorRowColor}{HTML}{DCEEDC}      %
\definecolor{vigorGainColor}{HTML}{6A9079}     %
\definecolor{vigorLossColor}{HTML}{B85C5C}     %
\definecolor{vigorZeroColor}{gray}{0.60}        %
\newcommand{\gainup}[1]{\,{\scriptsize\textcolor{vigorGainColor}{$\uparrow$#1}}}
\newcommand{\gaindn}[1]{\,{\scriptsize\textcolor{vigorGainColor}{$\downarrow$#1}}}
\newcommand{\lossdn}[1]{\,{\scriptsize\textcolor{vigorLossColor}{$\downarrow$#1}}}
\newcommand{\lossup}[1]{\,{\scriptsize\textcolor{vigorLossColor}{$\uparrow$#1}}}
\newcommand{\nochg}{}
\newcommand{\bestnum}[1]{%
  {\setlength{\fboxsep}{1pt}\colorbox{vigorRowColor}{\textbf{#1}}}%
  \kern-1pt%
}
\newcommand{\caplite}[1]{{\setlength{\fboxsep}{1pt}\colorbox{vigorRowColor}{#1}}}

\newtcolorbox{insightbox}{
  enhanced,
  colback=gray!6,
  colframe=black!45,
  boxrule=0.8pt,
  arc=4pt,
  before skip=8pt,
  after skip=8pt,
  left=6pt,
  right=6pt,
  top=7pt,
  bottom=7pt
}
\title{\modelname: Visual Gain Ordering for Hallucination Mitigation in Multimodal Discrete Diffusion Language Models}

\author{%
  Junzhe Chen$^{1}$\thanks{Equal contribution. \texttt{chenjz24@mails.tsinghua.edu.cn}, \texttt{tqin0515@uni.sydney.edu.au}.} \hspace{0.8em}%
  Tian Qin$^{1,2}$\footnotemark[1] \hspace{0.8em}%
  Eugenie Shi$^{3}$ \hspace{0.8em}%
  Tianshu Zhang$^{1}$ \hspace{0.8em}%
  Qiang Ju$^{4}$ \hspace{0.8em}%
  Lijie Wen$^{1}$\thanks{Corresponding author: \texttt{wenlj@tsinghua.edu.cn}.} \\[4pt]
  \normalfont
  $^{1}$Tsinghua University \quad
  $^{2}$The University of Sydney \quad
  $^{3}$Stanford University \quad
  $^{4}$Baichuan AI
}

\begin{document}

\maketitle

\begin{abstract}
Multimodal discrete diffusion language models (dLLMs) generate responses through iterative mask prediction and offer a promising alternative to autoregressive large vision-language models. However, they still suffer from hallucination, producing outputs that are plausible under language priors but insufficiently grounded in the image. We argue that hallucination in multimodal dLLMs is not only a token prediction problem, but also a \emph{premature commitment} problem: during iterative denoising, some masked positions are unmasked too early because their confidence is driven more by language priors than by visual evidence, and these early commitments can bias later generation. To address this issue, we propose \textbf{Vi}sual \textbf{G}ain \textbf{Or}dering (\modelname), a training-free and plug-and-play inference strategy that reorders masked positions according to counterfactual visual evidence. At each denoising step, \modelname compares the confidence of each candidate token under the original image-conditioned pass and an additional visual-ablation pass, and prioritizes positions whose predictions are more strongly supported by the image. Experiments on LLaDA-V and Lumina-DiMOO show that \modelname consistently reduces hallucination and improves multimodal reasoning performance across benchmarks. Code is available at \url{https://github.com/THU-BPM/Vigor}.

\end{abstract}

\section{Introduction}
\label{sec:intro}

Recent advances in large vision-language models have enabled strong performance on a wide range of multimodal tasks, including visual question answering, image-grounded dialogue, and multimodal reasoning~\cite{li2023blip2,liu2024improvedbaselinesvisualinstruction,Ye_2024,bai2025qwen25vltechnicalreport,geminiteam2025gemini}. Beyond conventional autoregressive LVLMs, recent work has begun to explore \emph{multimodal discrete diffusion language models} (dLLMs), which generate responses through iterative mask prediction rather than left-to-right decoding~\cite{He_2023,Zhou_2024,NEURIPS2024_be30024e,nie2025large,you2025lladavlargelanguagediffusion,xin2025luminadimooomnidiffusionlarge,yang2025mmadamultimodallargediffusion,li2026lavida}. By refining multiple positions in parallel, multimodal dLLMs offer a fundamentally different decoding paradigm with promising advantages in efficiency and controllability~\cite{wu2025fastdllmtrainingfreeaccelerationdiffusion,arriola2025encoderdecoder,xiong2026unveiling,zhu2025segment}. However, they still suffer from hallucination: the generated response may be plausible under language priors while remaining insufficiently grounded in the image~\cite{rohrbach2018object,li2023evaluating,Huang_2025,chang2026tracedet,guo2026lostdiffusion,10.1007/978-3-032-37422-6_19}.

\begin{figure}[t]
\centering
\includegraphics[width=\linewidth]{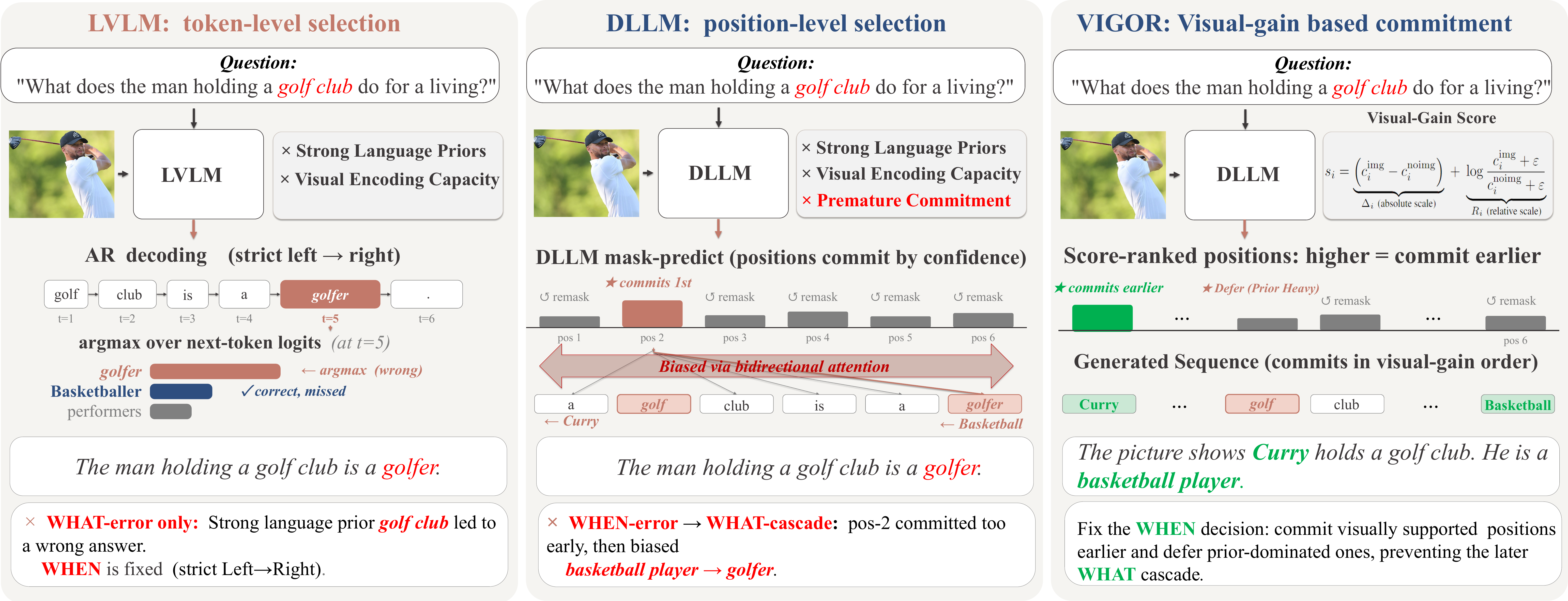}
\vspace{-4mm}
\caption{Comparison of hallucination mechanisms in autoregressive LVLMs, multimodal dLLMs, and \modelname.}
\label{fig:introduction}
\vspace{-8mm}
\end{figure}

Although hallucination has been widely studied in autoregressive LVLMs, the failure mode in multimodal dLLMs is importantly different. In autoregressive decoding, hallucination is typically viewed as a \emph{token selection} problem, where the model over-predicts linguistically plausible but visually unsupported tokens due to strong language priors or weak visual grounding~\cite{Tong_2024,Lee_2025,kaduri2025whats}. In multimodal dLLMs, by contrast, generation proceeds through iterative denoising, where the decoder must decide not only \emph{what} token to predict, but also \emph{when} each masked position should be committed. Existing blockwise decoders usually rank masked positions by confidence and unmask the top-ranked ones first~\cite{nie2025large,huang2026dont}. However, this confidence can be misleading in multimodal settings. As shown in Figure~\ref{fig:introduction}, an autoregressive LVLM may predict \emph{golfer} because the phrase ``golf club'' in the prompt induces an overly strong language prior. A multimodal dLLM can amplify this failure: the position corresponding to \emph{golf}, which is highly confident due to the same prior rather than visual evidence, may be committed too early and bias subsequent denoising toward \emph{golfer}. Therefore, hallucination in multimodal dLLMs is not merely a prediction error, but also a \emph{premature commitment} problem.

This difference makes existing hallucination mitigation strategies less suitable for multimodal dLLMs. Prior methods mainly fall into three categories: fine-tuning with additional supervision~\cite{Sun_2024,Jiang_2024,Yu_2024,Yu_2024_2,Zhang_2024_2,Chen_Zhang_Huang_Niu_Sun_Zhang_Zhou_Wen_2026}, perceptual enhancement with auxiliary visual signals~\cite{Chen_2024,Liu_2024_3,An_2025,Woo_2025,yin2025lifting,Favero_2024}, and inference-time contrastive decoding or logit correction~\cite{Li_2023,chuang2023dola,leng2024mitigating,huang2024opera,Wang_2024,NEURIPS2024_f1592b0d,Liu_2025,NEURIPS2024_37294f03,Zhu_2025,Yue_2024,qin2026reallyneedexternaltools}. While effective for autoregressive LVLMs, these methods mainly target \emph{which token} should be generated, but do not directly address \emph{which positions} should be committed earlier or later during iterative denoising. As a result, they are not designed for the distinctive hallucination mechanism of multimodal dLLMs.

To address this issue, we propose \textbf{Vi}sual \textbf{G}ain \textbf{Or}dering (\modelname), a training-free and plug-and-play inference strategy for multimodal discrete diffusion language models. Instead of modifying model parameters or redesigning the decoder, \modelname intervenes only in the commitment order of masked positions. The key idea is to prioritize positions whose current predictions are genuinely supported by the image. To estimate such support, \modelname introduces a counterfactual visual-ablation pass during decoding. At each denoising step, the model first performs the standard image-conditioned forward pass, and then an additional visual-ablation pass with the visual input ablated. By comparing the confidence of the same candidate token across the two passes, \modelname measures its \emph{visual gain}, i.e., how much the current prediction depends on the image rather than on language priors alone. As illustrated in Figure~\ref{fig:introduction}, this allows the model to commit visually grounded cues such as \emph{Curry} earlier, while deferring prior-dominated tokens such as \emph{golf}. As a result, subsequent denoising is steered toward the correct answer \emph{basketball player}, rather than being prematurely biased toward \emph{golfer}.

In this way, \modelname turns decoding in multimodal dLLMs into an \emph{evidence-grounded commitment} process. Rather than changing which token a position prefers, it changes when that position is allowed to become committed. Because \modelname only replaces the ordering criterion while keeping the original blockwise decoding pipeline intact, it is model-agnostic and requires no retraining. Extensive experiments on LLaDA-V and Lumina-DiMOO show that \modelname consistently reduces hallucination and improves multimodal reasoning performance across benchmarks.

We list our insights and our contributions are as follows:

\begin{insightbox}
\noindent{\large\textcolor{brown!55!black}{\faLightbulb[regular]}}\hspace{0.45em}\textit{In multimodal dLLMs, hallucination is not only about which token is chosen, but also about when positions are committed: prior-driven positions exposed too early can bias later denoising.}
\end{insightbox}

$\bullet$ We identify \emph{premature commitment} as a distinctive prior-driven
hallucination mechanism in multimodal discrete diffusion language models,
showing that hallucination arises not only from erroneous token
prediction, but also from unreliable confidence-based commitment order
during iterative denoising.

$\bullet$ We propose \textbf{Vi}sual \textbf{G}ain \textbf{Or}dering (\modelname), a novel training-free and plug-and-play inference strategy that introduces counterfactual visual evidence into blockwise decoding and reorders masked positions according to their visual support, without finetuning or architectural modification.

$\bullet$ Extensive experiments on LLaDA-V and Lumina-DiMOO across three hallucination benchmarks (POPE, CHAIR, and AMBER) demonstrate that \modelname consistently improves all evaluated metrics on both models (e.g., average POPE accuracy improves by $4.7$ and $2.6$ points and F1 by $3.1$ and $8.3$ points), validating its effectiveness, robustness, and model-agnostic generalizability.

\section{Related Work}
\label{sec:related_work}

\subsection{Multimodal Discrete Diffusion Models}
\label{sec:mdllm}

Discrete diffusion language models (dLLMs)~\cite{He_2023,Zhou_2024,NEURIPS2024_be30024e,NEURIPS2024_bad233b9,NEURIPS2024_eb0b13cc,nie2025large,yu2025discretediffusionlargelanguage} replace autoregressive, left-to-right decoding with iterative demasking under bidirectional attention. Recent work extends this paradigm to multimodal inputs: LLaDA-V~\cite{you2025lladavlargelanguagediffusion}, Lumina-DiMOO~\cite{xin2025luminadimooomnidiffusionlarge}, MMaDA~\cite{yang2025mmadamultimodallargediffusion}, and LaViDa~\cite{li2026lavida} couple a visual encoder with a dLLM through a lightweight projection module~\cite{li2023blip2,liu2024improvedbaselinesvisualinstruction,Ye_2024,bai2025qwen25vltechnicalreport,geminiteam2025gemini}, forming an emerging class of multimodal dLLMs (mdLLMs). A parallel line of work further improves the efficiency and controllability of dLLM decoding~\cite{wu2025fastdllmtrainingfreeaccelerationdiffusion,arriola2025encoderdecoder,huang2025ctrldiffboostinglargediffusion,zhu2025segment,xiong2026unveiling}.

The decoding mechanism of mdLLMs differs fundamentally from that of autoregressive LVLMs~\cite{touvron2023llama,openai2024gpt4technicalreport,geminiteam2025gemini}. Rather than generating tokens one at a time in left-to-right order, an mdLLM predicts candidate tokens for all currently masked positions in parallel and then commits a confidence-ranked subset at each denoising step. This confidence-based blockwise decoding is central to our setting: because the commitment order is decided online, positions dominated by language priors can receive high confidence and be exposed too early, even when they are not supported by the image~\cite{Tong_2024,Lee_2025,kaduri2025whats}. Once committed, such tokens become visible to the remaining masked positions through bidirectional attention, allowing hallucinated content to propagate and reinforce itself across later denoising steps.

\subsection{Mitigating Hallucinations in LVLMs}
\label{sec:hallucination}

A large body of work addresses hallucination in autoregressive LVLMs~\cite{rohrbach2018object,li2023evaluating,Gunjal_2024,jiang2024hal,Zhong_2024,Lee_2025,Xia_2025,Huang_2025}. Existing approaches fall into three categories: training-time supervision~\cite{Sun_2024,Jiang_2024,Yu_2024,Yu_2024_2,Zhang_2024_2,Chen_Zhang_Huang_Niu_Sun_Zhang_Zhou_Wen_2026}, perceptual enhancement via attention or activation intervention~\cite{Chen_2024,Liu_2024_3,An_2025,Woo_2025,yin2025lifting,Favero_2024,heimersheim2024useinterpretactivationpatching,chen2025ict,Yin_2024}, and inference-time contrastive or modified decoding~\cite{Li_2023,chuang2023dola,leng2024mitigating,huang2024opera,Wang_2024,NEURIPS2024_f1592b0d,Liu_2025,NEURIPS2024_37294f03,Zhu_2025,Yue_2024,qin2026reallyneedexternaltools}. Despite this diversity, all such methods target autoregressive generation, where token order is fixed and hallucination mitigation acts on token preference under a left-to-right prefix.

This assumption breaks down in multimodal dLLMs, where masked positions are updated in parallel and the commitment order is decided dynamically during denoising. A handful of concurrent works begin to analyze hallucination in dLLMs and explore remedies such as remasking~\cite{chang2026tracedet,guo2026lostdiffusion,huang2026dont}, but they target text-only dLLMs and leave the interaction between visual evidence and language priors in multimodal decoding unaddressed. Consequently, existing hallucination mitigation methods do not provide a native solution for multimodal dLLMs.

\section{Method}
\label{sec:method}

We present \textbf{Vi}sual \textbf{G}ain \textbf{Or}dering (\modelname). \modelname is a training-free, plug-and-play inference strategy for multimodal dLLMs that reorders the commitment of masked positions according to visual evidence. Rather than modifying model parameters or redesigning the decoder, \modelname intervenes only in \emph{which} positions are committed earlier during iterative denoising.

\subsection{Blockwise Decoding in Multimodal dLLMs}

We consider a multimodal dLLM that generates an answer sequence through iterative mask prediction. Given an image $I$ and a prompt $q$, the model maintains a partially masked response 
$\mathbf{z}^{(s)} = \{z^{(s)}_1,\ldots,z^{(s)}_T\}$ at denoising step $s$, where each position is either a committed token or a \texttt{[MASK]} token. Let $\mathcal{M}^{(s)}=\{i:z^{(s)}_i=\texttt{[MASK]}\}$ denote the set of currently masked positions. At each step, the model predicts a distribution over the vocabulary for every masked position:
\begin{equation}
    p^{(s)}_i = p_\theta \bigl(y_i \mid I, q, \mathbf{z}^{(s)} \bigr), 
    \qquad i \in \mathcal{M}^{(s)} .
\end{equation}

\begin{figure}[t]
    \centering
    \includegraphics[width=\linewidth,page=1]{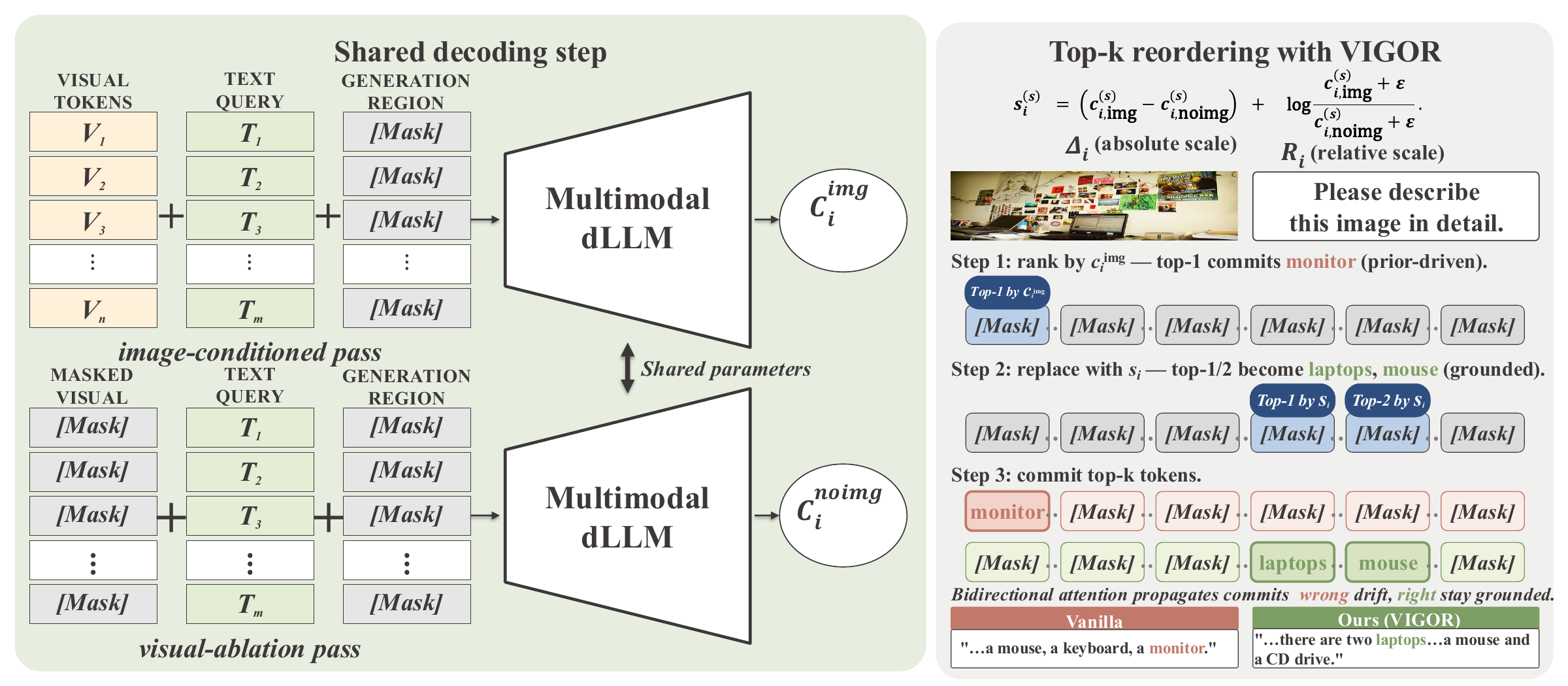}
    \vspace{-4mm}
    \caption{Overview of \modelname in blockwise decoding. At each decoding step, the model runs image-conditioned and visual-ablation passes with shared parameters, computes a \modelname score for every masked position, and selects the top-$k$ positions to unmask within the current block.}
    \vspace{-4mm}
    
    \label{fig:vigor_overview}
\end{figure}

The candidate token and its confidence are given by $\hat{y}^{(s)}_i\!=\!\arg\max_{y \in \mathcal{V}} p^{(s)}_i(y)$ and $c^{(s)}_i\!=\! p^{(s)}_i\bigl(\hat{y}^{(s)}_i\bigr)$, where $\mathcal{V}$ is the vocabulary. In standard blockwise decoding, the decoder ranks masked positions by $c^{(s)}_i$ and commits the top-$k_s$ positions in the active block: $\mathcal{C}^{(s)}
    =
    \operatorname{TopK}_{i \in \mathcal{M}^{(s)} \cap \mathcal{B}^{(s)}}
    \bigl(c^{(s)}_i, k_s\bigr)$, where $\mathcal{B}^{(s)}$ denotes the current decoding block and $k_s$ is determined by the decoding schedule. The partially generated sequence is updated as:
\begin{equation}
\setlength\abovedisplayskip{0mm} \setlength\belowdisplayskip{0mm}
    z^{(s+1)}_i =
    \begin{cases}
    \hat{y}^{(s)}_i, & i \in \mathcal{C}^{(s)}, \\
    z^{(s)}_i, & \text{otherwise}.
    \end{cases}
\end{equation}

This confidence-based ordering is unreliable in multimodal generation because the confidence $c^{(s)}_i$ does not distinguish visual evidence from language priors. Some positions may receive high confidence mainly because their candidate tokens are likely under the prompt and partially generated text, rather than because they are grounded in the image. Once such a prior-dominated position $i$ is committed early, its token $\hat{y}^{(s)}_i$ becomes visible to all remaining masked positions through bidirectional attention. Consequently, later predictions are conditioned on this early commitment,
\begin{equation}
    p^{(s+1)}_j
    =
    p_\theta \bigl(y_j \mid I, q, \mathbf{z}^{(s+1)} \bigr),
    \qquad j \in \mathcal{M}^{(s+1)},
\end{equation}
which allows a hallucinated structure to propagate and reinforce itself in subsequent denoising steps. \modelname replaces only the ordering score used to select $\mathcal{C}^{(s)}$, prioritizing masked positions by visual support rather than raw confidence alone.

\subsection{Counterfactual Visual Ablation}

The standard confidence $c_i^{(s)}$ in blockwise decoding does not reveal whether a candidate token is supported by the image or merely by language priors. To estimate this dependency, \modelname introduces a counterfactual visual-ablation pass at each denoising step. In addition to the standard image-conditioned distribution
\begin{equation}
    p_{i,\mathrm{img}}^{(s)}
    =
    p_\theta \bigl(y_i \mid I, q, \mathbf{z}^{(s)} \bigr),
    \qquad i \in \mathcal{M}^{(s)},
\end{equation}
we run a second forward pass with the visual input ablated while keeping the text context and the partially generated response unchanged:
\begin{equation}
    p_{i,\mathrm{noimg}}^{(s)}
    =
    p_\theta \bigl(y_i \mid \varnothing, q, \mathbf{z}^{(s)} \bigr),
    \qquad i \in \mathcal{M}^{(s)}.
\end{equation}
Here $\varnothing$ denotes the masked or removed visual input. Importantly, the visual-ablation pass is not used to choose a new token. Instead, it provides a counterfactual confidence for the same candidate token selected by the image-conditioned pass. If this candidate remains highly confident after visual ablation, its preference is likely dominated by language priors; if its confidence drops substantially, the candidate is more likely to be visually grounded. This contrast forms the basis of \modelname.

\subsection{Visual Gain Ordering}

As shown in Figure \ref{fig:vigor_overview}, for each masked position $i$ at denoising step $s$, let
$\hat{y}_i^{(s)}=\arg\max_{y \in \mathcal{V}} p_{i,\mathrm{img}}^{(s)}(y)$, 
denote the candidate token selected by the image-conditioned pass. We then evaluate the confidence of this same candidate under the image-conditioned and visual-ablation distributions:
\begin{equation}
    c_{i,\mathrm{img}}^{(s)}
    =
    p_{i,\mathrm{img}}^{(s)}\bigl(\hat{y}_i^{(s)}\bigr),
    \qquad
    c_{i,\mathrm{noimg}}^{(s)}
    =
    p_{i,\mathrm{noimg}}^{(s)}\bigl(\hat{y}_i^{(s)}\bigr).
\end{equation}
Since we cannot directly observe whether a candidate token is visually grounded during decoding, we use visual ablation as a local counterfactual test. A useful ordering score should reward candidates that are confident with the image, while penalizing candidates that remain confident even without the image. We instantiate this criterion with a parameter-free visual-gain score:
\begin{equation}
\setlength\abovedisplayskip{0mm} \setlength\belowdisplayskip{0mm}
\label{eq:vigor_score}
s_i^{(s)}
=
\underbrace{\left(c_{i,\mathrm{img}}^{(s)} - c_{i,\mathrm{noimg}}^{(s)}\right)}_{\Delta_i^{(s)}\ \text{(absolute scale)}}
\;+\;
\underbrace{\log \frac{c_{i,\mathrm{img}}^{(s)} + \varepsilon}{c_{i,\mathrm{noimg}}^{(s)} + \varepsilon}}_{R_i^{(s)}\ \text{(relative scale)}} .
\end{equation}

The absolute term $\Delta_i^{(s)}$ measures how much the image increases the candidate confidence, preventing the score from favoring tokens that have a large ratio but low absolute confidence. The relative term $R_i^{(s)}$ measures how much more likely the candidate is with the image than without it, with $\varepsilon=10^{-8}$ for numerical stability. These two terms are complementary: the absolute scale suppresses low-confidence noise, while the relative scale highlights candidates whose confidence is specifically induced by visual evidence.

At each decoding step, \modelname replaces the raw confidence score in standard blockwise decoding with $s_i^{(s)}$. The committed positions are therefore selected as $\mathcal{C}^{(s)}=\operatorname{TopK}_{i \in \mathcal{M}^{(s)} \cap \mathcal{B}^{(s)}}
    \bigl(s_i^{(s)}, k_s\bigr)$, while the candidate token $\hat{y}_i^{(s)}$ itself remains unchanged. In this sense, \modelname does not alter \emph{which} token a position currently prefers; it changes \emph{when} that position is allowed to commit. Positions whose predictions depend more strongly on the image are committed earlier, whereas prior-dominated positions are deferred. This schedule-level intervention keeps the denoising trajectory aligned with visual evidence and reduces structural hallucinations caused by early prior-driven commitments. Because \modelname only changes the ordering score, it requires no finetuning, extra supervision, or architectural modification.

\section{Experiments}
\label{sec:experiments}

\subsection{Experimental Settings}
\label{sec:exp-setup}

We evaluate \modelname on two recent multimodal discrete diffusion language models and three hallucination-sensitive benchmarks that together cover yes/no object probing, open-ended caption hallucination, and mixed generative--discriminative grounding.

\noindent \textbf{Datasets and metrics.} We report results on POPE \citep{li2023evaluating}, CHAIR \citep{rohrbach2018object}, and AMBER \citep{wang2023amber}. POPE is a benchmark for assessing whether LVLMs can correctly recognize the presence of specific objects in images, thereby exposing object-level hallucinations. It uses balanced yes/no questions with random, popular, and adversarial sampling strategies, and draws from MSCOCO \citep{lin2014microsoft}, A-OKVQA \citep{schwenk2022okvqa}, and GQA \citep{hudson2019gqa}. Following the standard setup, we report accuracy and F1 on all nine subsets. CHAIR evaluates hallucination in open-ended image captioning. We follow the common setting based on 500 randomly sampled images from COCO 2014 val \citep{lin2014microsoft} and report CHAIR$_S$ and CHAIR$_I$, where lower values indicate fewer hallucinated objects. AMBER provides a broader benchmark for hallucination and grounding quality through both generative and discriminative tasks. For the generative part, we report CHAIR, Cover, Hal, and Cog; for the discriminative part, we report Acc and F1.

\noindent \textbf{Baseline models.} We evaluate \modelname on two state-of-the-art multimodal discrete diffusion language models, LLaDA-V \citep{you2025lladavlargelanguagediffusion} and Lumina-DiMOO \citep{xin2025luminadimooomnidiffusionlarge}. For each model, we compare \modelname with vanilla decoding and two representative inference-time hallucination mitigation baselines, ICT and VCD. Details of how ICT~\cite{chen2025ict} and VCD~\cite{leng2024mitigating} are adapted to multimodal dLLM decoding are provided in Appendix~\ref{app:baselines}. All methods keep the underlying weights and prompt format unchanged, thereby isolating the effect of inference-time decoding interventions.

\noindent \textbf{Implementation details.} \modelname is applied only at inference time. At each decoding step, we run the standard image-conditioned forward pass together with a visual-ablation pass, compute the ordering score in Eq.~\ref{eq:vigor_score}, and use it to select the top-$k$ masked positions inside the current block. We keep the generation length, block schedule, and base model parameters unchanged and only replace the ranking rule used for blockwise commitment. The baseline and \modelname share the same blockwise decoding configuration. All experiments use greedy decoding (temperature $=0$), so the baseline and \modelname differ \emph{only} in the ranking score $s_i$ and the runs are deterministic given fixed inputs. All experiments are conducted on 3 NVIDIA A100 GPUs.

\subsection{Results on POPE}
Table~\ref{tab:pope_main} presents the results of LLaDA-V and Lumina-DiMOO on nine subsets of the POPE dataset. In addition to the vanilla baseline, we compare with two representative inference-time hallucination mitigation methods originally proposed for autoregressive LVLMs, VCD~\cite{leng2024mitigating} and ICT~\cite{chen2025ict}, which we adapt to multimodal dLLM decoding by applying their logit-correction rule at every denoising step (see Appendix~\ref{app:baselines} for how these baselines are adapted to mask-diffusion decoding). From these results, we make the following observations. 

\textbf{1)} \modelname consistently outperforms the vanilla baseline as well as the VCD and ICT baselines on every subset and every model, demonstrating that visual-gain-based reordering is effective even for simple yes/no object grounding. On average, \modelname improves accuracy by $4.7$ points and F1 by $3.1$ points over the vanilla baseline for LLaDA-V, and improves accuracy by $2.6$ points and F1 by $8.3$ points for Lumina-DiMOO. The gains over the next-best inference-time baseline (VCD or ICT) are also consistent across all nine subsets, indicating that the benefit of \modelname comes from the commitment-order intervention itself rather than from generic logit-level correction. \textbf{2)} The improvements are more pronounced on the popular and adversarial splits, where language-prior-driven mistakes are more likely to occur. For example, on the COCO popular split, F1 increases from $76.8$ to $80.5$ for LLaDA-V and from $66.8$ to $79.3$ for Lumina-DiMOO; on the COCO adversarial split, F1 improves from $76.3$ to $78.9$ and from $65.1$ to $74.8$, respectively. Similar trends are observed on A-OKVQA and GQA, suggesting that \modelname brings larger benefits when the benchmark places stronger pressure on language priors. \textbf{3)} The two models show different gain profiles: LLaDA-V obtains larger average gains in accuracy, while Lumina-DiMOO benefits more substantially in F1. We attribute this difference to model-dependent calibration, but the consistent improvements across all subsets confirm the robustness and model-agnostic effectiveness of \modelname.

\subsection{CHAIR Results}

Table~\ref{tab:chair_amber_main} summarizes the CHAIR results for open-ended caption generation at two output lengths. We follow the standard definitions:
\[
\mathrm{CHAIR}_S = \frac{|\text{captions with hallucinated objects}|}{|\text{all captions}|},
\qquad
\mathrm{CHAIR}_I = \frac{|\text{hallucinated objects}|}{|\text{all mentioned objects}|}.
\]
From these results, we make the following observations. 

\textbf{1)} \modelname consistently reduces caption hallucination for both models at both output lengths. Averaged over lengths $64$ and $512$, \modelname reduces CHAIR$_S$ by $6.50$ points for both LLaDA-V and Lumina-DiMOO, and reduces CHAIR$_I$ by $4.05$ points on LLaDA-V and $7.40$ points on Lumina-DiMOO. This shows that the benefit of visual-gain-based reordering extends beyond yes/no object probing to open-ended generation. \textbf{2)} The improvements are especially pronounced at length $64$. For LLaDA-V, CHAIR$_S$ drops from $32.7$ to $24.8$ and CHAIR$_I$ from $22.1$ to $18.8$; for Lumina-DiMOO, CHAIR$_S$ decreases from $33.3$ to $26.1$, while CHAIR$_I$ drops more sharply from $24.5$ to $14.4$. At length $512$, both models still improve on both metrics, although the margins become smaller. This length-dependent trend suggests that \modelname is most effective when early commitment decisions have a stronger influence on the final caption. \textbf{3)} Lumina-DiMOO obtains a larger reduction on CHAIR$_I$, especially at length $64$, while LLaDA-V also shows stable gains across both lengths. This indicates that once the commitment order is made more visually grounded, hallucinated object mentions are less likely to be introduced early and survive into the final caption.
\begin{table}[h]
    \centering
    \footnotesize
    \vspace{-2mm}
    \caption{Main results on POPE. We compare \modelname against the vanilla \emph{Baseline}, VCD and ICT. The values highlighted in \caplite{green} indicate the best metric under each setting. Arrows next to each value report the change vs.\ \emph{Baseline} ($\uparrow$/$\downarrow$ = improvement on higher-/lower-is-better metrics).}
    \vspace{-2mm}
    
    \label{tab:pope_main}
    \setlength{\tabcolsep}{5pt}
    \renewcommand{\arraystretch}{0.8}
    \begin{tabular*}{\linewidth}{@{\extracolsep{\fill}}lllllll@{}}
        \toprule
        \multirow{2}{*}{Dataset} & \multirow{2}{*}{Setting} & \multirow{2}{*}{Method}
          & \multicolumn{2}{c}{LLaDA-V} & \multicolumn{2}{c}{Lumina-DiMOO} \\
        \cmidrule(lr){4-5} \cmidrule(l){6-7}
         & & & \multicolumn{1}{c}{Accuracy$\uparrow$} & \multicolumn{1}{c}{F1$\uparrow$}
             & \multicolumn{1}{c}{Accuracy$\uparrow$} & \multicolumn{1}{c}{F1$\uparrow$} \\
        \midrule
        \multirow{12}{*}{COCO} & \multirow{4}{*}{Random}
          & Baseline   & 83.7\nochg          & 85.1\nochg          & 75.1\nochg          & 67.7\nochg \\
        & & VCD        & 84.5\gainup{0.8}    & 85.9\gainup{0.8}    & 76.0\gainup{0.9}    & 68.4\gainup{0.7} \\
        & & ICT        & 85.1\gainup{1.4}    & 86.8\gainup{1.7}    & 76.7\gainup{1.6}    & 70.1\gainup{2.4} \\
        & & \modelname & \bestnum{85.9}\gainup{2.2} & \bestnum{87.6}\gainup{2.5} & \bestnum{80.4}\gainup{5.3} & \bestnum{79.3}\gainup{11.6} \\
        \cmidrule(lr){2-7}
        & \multirow{4}{*}{Popular}
          & Baseline   & 71.9\nochg          & 76.8\nochg          & 74.1\nochg          & 66.8\nochg \\
        & & VCD        & 74.6\gainup{2.7}    & 78.9\gainup{2.1}    & 75.6\gainup{1.5}    & 67.7\gainup{0.9} \\
        & & ICT        & 72.6\gainup{0.7}    & 76.7\lossdn{0.1}    & 75.2\gainup{1.1}    & 68.1\gainup{1.3} \\
        & & \modelname & \bestnum{76.9}\gainup{5.0} & \bestnum{80.5}\gainup{3.7} & \bestnum{80.3}\gainup{6.2} & \bestnum{79.3}\gainup{12.5} \\
        \cmidrule(lr){2-7}
        & \multirow{4}{*}{Adversarial}
          & Baseline   & 71.1\nochg          & 76.3\nochg          & 72.2\nochg          & 65.1\nochg \\
        & & VCD        & 73.1\gainup{2.0}    & 77.9\gainup{1.6}    & 73.4\gainup{1.2}    & 65.4\gainup{0.3} \\
        & & ICT        & 70.2\lossdn{0.9}    & 76.0\lossdn{0.3}    & 72.7\gainup{0.5}    & 66.2\gainup{1.1} \\
        & & \modelname & \bestnum{74.7}\gainup{3.6} & \bestnum{78.9}\gainup{2.6} & \bestnum{74.9}\gainup{2.7} & \bestnum{74.8}\gainup{9.7} \\
        \midrule
        \multirow{12}{*}{A-OKVQA} & \multirow{4}{*}{Random}
          & Baseline   & 78.7\nochg          & 82.1\nochg          & 79.0\nochg          & 75.0\nochg \\
        & & VCD        & 81.1\gainup{2.4}    & 83.6\gainup{1.5}    & 79.5\gainup{0.5}    & 74.8\lossdn{0.2} \\
        & & ICT        & 79.4\gainup{0.7}    & 81.9\lossdn{0.2}    & 79.2\gainup{0.2}    & 75.8\gainup{0.8} \\
        & & \modelname & \bestnum{81.9}\gainup{3.2} & \bestnum{85.0}\gainup{2.9} & \bestnum{80.0}\gainup{1.0} & \bestnum{80.1}\gainup{5.1} \\
        \cmidrule(lr){2-7}
        & \multirow{4}{*}{Popular}
          & Baseline   & 66.8\nochg          & 74.6\nochg          & 77.3\nochg          & 73.5\nochg \\
        & & VCD        & 71.0\gainup{4.2}    & 76.6\gainup{2.0}    & 78.4\gainup{1.1}    & 73.7\gainup{0.2} \\
        & & ICT        & 67.8\gainup{1.0}    & 74.1\lossdn{0.5}    & 77.9\gainup{0.6}    & 74.0\gainup{0.5} \\
        & & \modelname & \bestnum{72.4}\gainup{5.6} & \bestnum{77.9}\gainup{3.3} & \bestnum{79.3}\gainup{2.0} & \bestnum{79.7}\gainup{6.2} \\
        \cmidrule(lr){2-7}
        & \multirow{4}{*}{Adversarial}
          & Baseline   & 61.1\nochg          & 71.4\nochg          & 69.3\nochg          & 65.6\nochg \\
        & & VCD        & 64.6\gainup{3.5}    & 73.3\gainup{1.9}    & 70.4\gainup{1.1}    & 65.9\gainup{0.3} \\
        & & ICT        & 60.8\lossdn{0.3}    & 71.8\gainup{0.4}    & 69.6\gainup{0.3}    & 67.1\gainup{1.5} \\
        & & \modelname & \bestnum{65.7}\gainup{4.6} & \bestnum{73.8}\gainup{2.4} & \bestnum{71.6}\gainup{2.3} & \bestnum{73.9}\gainup{8.3} \\
        \midrule
        \multirow{12}{*}{GQA} & \multirow{4}{*}{Random}
          & Baseline   & 70.1\nochg          & 76.9\nochg          & 77.2\nochg          & 72.1\nochg \\
        & & VCD        & 76.5\gainup{6.4}    & 80.9\gainup{4.0}    & 77.9\gainup{0.7}    & 72.4\gainup{0.3} \\
        & & ICT        & 69.6\lossdn{0.5}    & 76.4\lossdn{0.5}    & 77.8\gainup{0.6}    & 72.9\gainup{0.8} \\
        & & \modelname & \bestnum{78.1}\gainup{8.0} & \bestnum{82.2}\gainup{5.3} & \bestnum{79.6}\gainup{2.4} & \bestnum{79.8}\gainup{7.7} \\
        \cmidrule(lr){2-7}
        & \multirow{4}{*}{Popular}
          & Baseline   & 62.6\nochg          & 72.7\nochg          & 73.1\nochg          & 68.6\nochg \\
        & & VCD        & 67.4\gainup{4.8}    & 74.9\gainup{2.2}    & 74.0\gainup{0.9}    & 68.5\lossdn{0.1} \\
        & & ICT        & 63.2\gainup{0.6}    & 72.0\lossdn{0.7}    & 72.9\lossdn{0.2}    & 69.1\gainup{0.5} \\
        & & \modelname & \bestnum{69.0}\gainup{6.4} & \bestnum{76.2}\gainup{3.5} & \bestnum{75.0}\gainup{1.9} & \bestnum{76.3}\gainup{7.7} \\
        \cmidrule(lr){2-7}
        & \multirow{4}{*}{Adversarial}
          & Baseline   & 58.1\nochg          & 70.3\nochg          & 71.7\nochg          & 67.5\nochg \\
        & & VCD        & 62.0\gainup{3.9}    & 72.4\gainup{2.1}    & 71.8\gainup{0.1}    & 67.6\gainup{0.1} \\
        & & ICT        & 58.4\gainup{0.3}    & 70.5\gainup{0.2}    & 71.4\lossdn{0.3}    & 68.2\gainup{0.7} \\
        & & \modelname & \bestnum{62.3}\gainup{4.2} & \bestnum{72.4}\gainup{2.1} & \bestnum{71.9}\gainup{0.2} & \bestnum{74.2}\gainup{6.7} \\
        \bottomrule
    \end{tabular*}
    \vspace{-4mm}
\end{table}

\begin{table}[h]
\centering
\footnotesize
\vspace{-2mm}
\caption{Main results on CHAIR and AMBER. CHAIR reports C$_S$/C$_I$ at two caption lengths (64, 512); AMBER includes a generative track (\textbf{G}: CHAIR, Cover, Hal, Cog) and a discriminative track (\textbf{D}: Acc, F1). The values highlighted in \caplite{green} indicate the best metric for each model.}
\vspace{-2mm}

\label{tab:chair_amber_main}

\setlength{\tabcolsep}{4pt}

\resizebox{\linewidth}{!}{%
\begin{tabular}{@{}ll*{10}{l}@{}}
\toprule
\multirow{2}{*}{Model} & \multirow{2}{*}{Method}
  & \multicolumn{2}{c}{CHAIR Len 64}
  & \multicolumn{2}{c}{CHAIR Len 512}
  & \multicolumn{4}{c}{AMBER G}
  & \multicolumn{2}{c}{AMBER D} \\
\cmidrule(lr){3-4}\cmidrule(lr){5-6}\cmidrule(lr){7-10}\cmidrule(l){11-12}
  & & \multicolumn{1}{c}{C$_S\!\!\downarrow$} & \multicolumn{1}{c}{C$_I\!\!\downarrow$}
  & \multicolumn{1}{c}{C$_S\!\!\downarrow$} & \multicolumn{1}{c}{C$_I\!\!\downarrow$}
  & \multicolumn{1}{c}{CHAIR$\downarrow$} & \multicolumn{1}{c}{Cover$\uparrow$}
  & \multicolumn{1}{c}{Hal$\downarrow$} & \multicolumn{1}{c}{Cog$\downarrow$}
  & \multicolumn{1}{c}{Acc$\uparrow$} & \multicolumn{1}{c}{F1$\uparrow$} \\
\midrule
\multirow{2}{*}{LLaDA-V}
  & Baseline   & 32.7\nochg & 22.1\nochg & 47.5\nochg & 31.0\nochg
                & 6.0\nochg  & 35.4\nochg & 21.1\nochg & 1.2\nochg
                & 73.3\nochg & 81.7\nochg \\
  & \modelname & \bestnum{24.8}\gaindn{7.9}  & \bestnum{18.8}\gaindn{3.3}
                & \bestnum{42.4}\gaindn{5.1} & \bestnum{26.2}\gaindn{4.8}
                & \bestnum{5.0}\gaindn{1.0}  & \bestnum{39.8}\gainup{4.4}
                & \bestnum{16.9}\gaindn{4.2} & \bestnum{0.6}\gaindn{0.6}
                & \bestnum{77.1}\gainup{3.8} & \bestnum{84.1}\gainup{2.4} \\
\addlinespace[1pt]
\multirow{2}{*}{Lumina-DiMOO}
  & Baseline   & 33.3\nochg & 24.5\nochg & 49.1\nochg & 30.7\nochg
                & 20.1\nochg & 47.3\nochg & 67.5\nochg & 9.5\nochg
                & 67.4\nochg & 76.8\nochg \\
  & \modelname & \bestnum{26.1}\gaindn{7.2}  & \bestnum{14.4}\gaindn{10.1}
                & \bestnum{43.3}\gaindn{5.8} & \bestnum{26.0}\gaindn{4.7}
                & \bestnum{16.7}\gaindn{3.4} & \bestnum{49.8}\gainup{2.5}
                & \bestnum{50.2}\gaindn{17.3}& \bestnum{3.8}\gaindn{5.7}
                & \bestnum{70.8}\gainup{3.4} & \bestnum{78.0}\gainup{1.2} \\
\bottomrule
\end{tabular}%
}
\vspace{-4mm}

\end{table}

\subsection{AMBER Results}
Table~\ref{tab:chair_amber_main} further evaluates \modelname on AMBER, which includes both generative (G) and discriminative (D) hallucination-related metrics. From the results, we draw the following conclusions. 

\textbf{1)} \modelname improves all AMBER metrics for both models. On LLaDA-V, the generative CHAIR score decreases from $6.0$ to $5.0$, Cover increases from $35.4$ to $39.8$, and the discriminative metrics also improve, with Acc rising from $73.3$ to $77.1$ and F1 from $81.7$ to $84.1$. Lumina-DiMOO follows the same trend: CHAIR decreases from $20.1$ to $16.7$, Cover increases from $47.3$ to $49.8$, Acc improves from $67.4$ to $70.8$, and F1 improves from $76.8$ to $78.0$. \textbf{2)} The gains are particularly clear on hallucination-sensitive generative metrics. For Lumina-DiMOO, Hal drops substantially from $67.5$ to $50.2$, and Cog decreases from $9.5$ to $3.8$. Since Lumina-DiMOO starts from a weaker baseline with more severe hallucination, the larger improvement suggests that \modelname is especially helpful when prior-driven commitments are more frequent. \textbf{3)} The consistent improvement across both generative and discriminative metrics indicates that \modelname is not limited to caption-style outputs. Instead, by committing visually supported positions earlier and deferring prior-dominated ones, \modelname improves the overall grounding behavior of multimodal dLLMs during inference.

\section{Analysis}
\label{sec:analysis}

\subsection{Score Decomposition}
Table~\ref{tab:score_ablation} reports a decomposition ablation on LLaDA-V, COCO POPE in which the \modelname score is replaced with each of its two scales in turn: \textsc{Diff-only} retains only the absolute scale $\Delta_i$, and \textsc{LogRatio-only} retains only the relative scale $R_i$ defined in Eq.~\eqref{eq:vigor_score}. On the random split, both single-scale variants already improve over the baseline, so a single-scale score superficially appears sufficient. The adversarial split reverses this reading: \textsc{Diff-only} retains only about half of the full gain, while \textsc{LogRatio-only} drops $4.2$ acc \emph{below} the baseline and inflates the yes-rate from $72.2\%$ to $75.8\%$. The two failures land in the two regimes each scale is blind to. When $c^{\mathrm{img}}_i \approx c^{\mathrm{noimg}}_i$---the regime flooded by adversarial prompts that elevate both conditionals---$\Delta_i$ loses its signal while $R_i$ still separates image-supported tokens; when both conditionals approach the prior floor, $R_i$ amplifies tiny denominators into arbitrary scores while $\Delta_i$ correctly filters noise. Neither regime is visible on random, which is why single-scale scores look acceptable there; adversarial hits both regimes at once, and only the unified $s_i$ covers both, consistent with moving the yes-rate toward the balanced prior $50\%$ on both splits ($-3.9$ pts on random, $-4.9$ pts on adversarial).

\begin{table}[h]
    \centering
    \footnotesize
    \vspace{-2mm}
    \caption{Score decomposition ablation on LLaDA-V, COCO POPE. Yes\% values closer to the balanced prior $50\%$ indicate less object-bias. Best per column is highlighted in \caplite{green}.}
    \label{tab:score_ablation}
    \vspace{-2mm}
    \setlength{\tabcolsep}{6pt}
    \begin{tabular*}{\linewidth}{@{\extracolsep{\fill}}lllllll@{}}
        \toprule
        \multirow{2}{*}{Method} & \multicolumn{3}{c}{\textbf{Random}} & \multicolumn{3}{c}{\textbf{Adversarial}} \\
        \cmidrule(lr){2-4} \cmidrule(l){5-7}
         & \multicolumn{1}{c}{Acc$\uparrow$} & \multicolumn{1}{c}{F1$\uparrow$} & \multicolumn{1}{c}{Yes\%$\!\to\!50$}
         & \multicolumn{1}{c}{Acc$\uparrow$} & \multicolumn{1}{c}{F1$\uparrow$} & \multicolumn{1}{c}{Yes\%$\!\to\!50$} \\
        \midrule
        Baseline               & 83.7\nochg          & 85.1\nochg          & 59.6\nochg          & 71.1\nochg          & 76.3\nochg          & 72.2\nochg \\
        \textsc{Diff-only}     & 84.6\gainup{0.9}    & 85.8\gainup{0.7}    & 58.4\gaindn{1.2}    & 73.0\gainup{1.9}    & 77.0\gainup{0.7}    & 67.4\gaindn{4.8} \\
        \textsc{LogRatio-only} & 84.2\gainup{0.5}    & 85.5\gainup{0.4}    & 59.0\gaindn{0.6}    & 66.9\lossdn{4.2}    & 73.7\lossdn{2.6}    & 75.8\lossup{3.6} \\
        \modelname (full)      & \bestnum{85.9}\gainup{2.2} & \bestnum{87.6}\gainup{2.5} & \bestnum{55.7}\gaindn{3.9} & \bestnum{74.7}\gainup{3.6} & \bestnum{78.9}\gainup{2.6} & \bestnum{67.3}\gaindn{4.9} \\
        \bottomrule
    \end{tabular*}
    \vspace{-2mm}
\end{table}

\subsection{Commitment Reordering}
\begin{figure}[t]
    \centering
    \includegraphics[width=\linewidth]{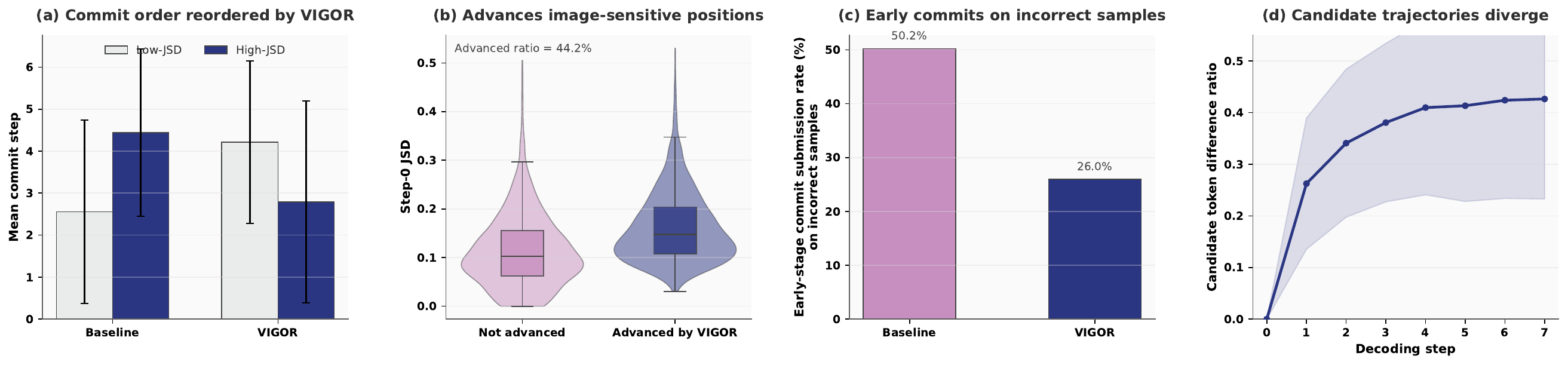}
    \vspace{-5mm}
    \caption{Diagnostic analysis on LLaDA-V over the AMBER benchmark. (a) Mean commit step for high- and low-JSD positions under Baseline and \modelname. (b) Step-$0$ JSD for positions advanced versus not advanced by \modelname. (c) Early-stage unstable commit rate on incorrect samples. (d) Candidate-token difference ratio across decoding steps.}
    \vspace{-4mm}
    \label{fig:discussion_analysis}
\end{figure}

Figure~\ref{fig:discussion_analysis} probes whether \modelname changes commitment order in the intended direction. \modelname does not change the generator itself; it changes \emph{when} each masked position is allowed to become visible to subsequent denoising steps. To measure visual sensitivity, we use the step-$0$ Jensen--Shannon divergence (JSD) between the image-conditioned and no-image token distributions:
\[
\mathrm{JSD}(p_i^{\mathrm{img}}, p_i^{\mathrm{noimg}})
=
\frac{1}{2}\mathrm{KL}(p_i^{\mathrm{img}}\|m_i)
+
\frac{1}{2}\mathrm{KL}(p_i^{\mathrm{noimg}}\|m_i),
\quad
m_i=\frac{1}{2}(p_i^{\mathrm{img}}+p_i^{\mathrm{noimg}}).
\]
A larger JSD indicates stronger sensitivity to visual ablation. Figure~\ref{fig:discussion_analysis}(a) shows that the baseline tends to commit low-JSD positions earlier, whereas \modelname shifts earlier commitments toward high-JSD positions. Figure~\ref{fig:discussion_analysis}(b) further shows that positions advanced by \modelname have higher step-$0$ JSD than those not advanced, with $44.2\%$ of analyzed positions committed earlier under \modelname. Figure~\ref{fig:discussion_analysis}(d) shows that this reordering propagates through decoding: the candidate-token difference ratio grows across decoding steps. Together, these results indicate that \modelname intervenes at the commitment-order level, exposing visually sensitive positions earlier and deferring less image-dependent ones.

\subsection{Cascade Suppression}
The main experiments show that \modelname consistently reduces hallucination. Figure~\ref{fig:discussion_analysis}(c) helps explain why: on incorrect samples in the LLaDA-V AMBER analysis, the early-stage unstable commit rate drops from $50.2\%$ under the baseline to $26.0\%$ under \modelname. This matters in multimodal dLLMs because decoding uses bidirectional attention: once an unstable position is committed too early, it becomes visible to the remaining masked positions through bidirectional attention, cascading the error into subsequent denoising steps. Reducing such early hallucinated anchors therefore prevents later error propagation. Combined with Figures~\ref{fig:discussion_analysis}(a) and \ref{fig:discussion_analysis}(b), this supports a single mechanism: \modelname advances positions that are more sensitive to visual evidence and suppresses unstable early commitments before they spread through later decoding.

\subsection{Case Study and Error Analysis}

\begin{figure}[t]
\centering
\includegraphics[width=\linewidth]{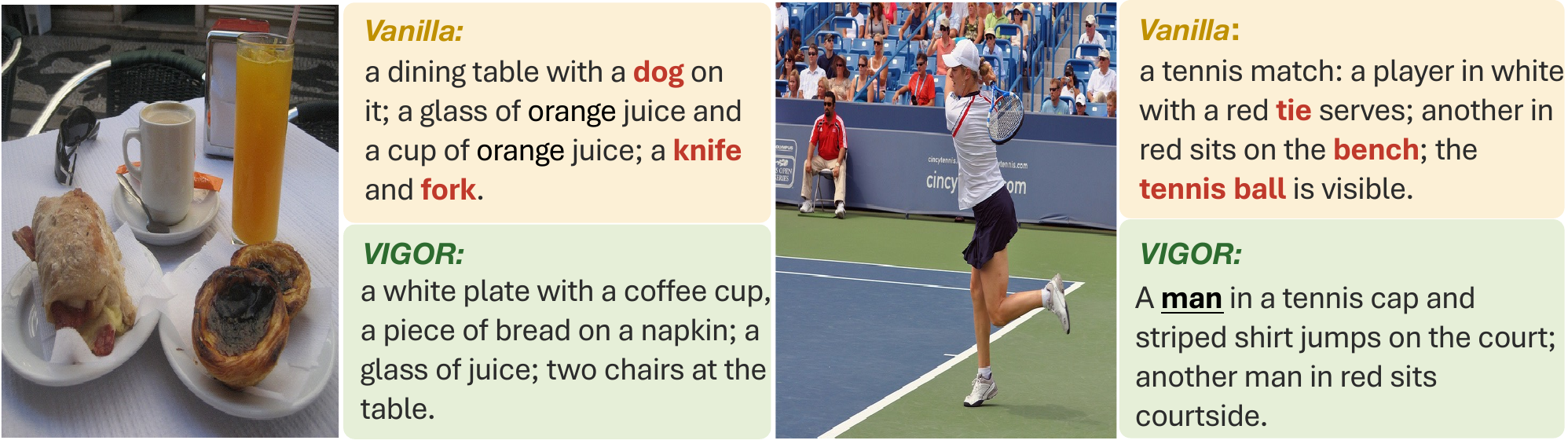}
\vspace{-5mm}
\caption{Case study and error analysis.}
\vspace{-5mm}

\label{fig:case_study_error}
\end{figure}

Figure~\ref{fig:case_study_error} compares vanilla decoding and \modelname on two examples, with hallucinated tokens highlighted in red. In the left case, vanilla decoding introduces several prior-driven dining co-occurrences, such as a dog, an extra cup of orange juice, and a knife and fork, none of which are supported by the image. In contrast, \modelname produces a more grounded description by correctly mentioning the coffee cup, bread, juice, and background chairs. In the right case, vanilla decoding hallucinates template-like tennis-scene details, including a red tie, a bench, and a visible tennis ball. \modelname removes these unsupported details and better captures the cap, striped shirt, jumping pose, and person on the side. However, it still misidentifies the player’s gender, suggesting that \modelname mitigates prior-driven hallucinations but cannot correct recognition errors inherited from the base model.

\section{Conclusion and Limitations}
\label{sec:conclusion}

We argue that hallucination in multimodal discrete diffusion language models is not only a token-prediction error but also a \emph{premature commitment} problem. To address this, we propose \modelname, a training-free plug-and-play inference method that controls \emph{when} masked positions commit during iterative denoising. By comparing the confidence of the same candidate token under image-conditioned and visual-ablation passes, \modelname prioritizes positions supported by visual evidence and defers those dominated by language priors. Experiments on LLaDA-V and Lumina-DiMOO across benchmarks show that changing the commitment order alone consistently reduces hallucination and improves multimodal reasoning. Diagnostic analysis further confirms that \modelname advances image-sensitive positions, defers less visually dependent ones, and suppresses unstable early commitments.

\textbf{Limitations.} \modelname requires an additional no-image forward pass at each denoising step, roughly doubling decoding time in our experiments. It also requires access to internal decoding scores and control over the commitment schedule, making it most suitable for open-weight or controllable models. Future work will explore more efficient variants and broader model settings.

\begin{ack}
This work is supported by the National Key Research and
Development Program of China (No.2024YFB3309702),
the National Nature Science Foundation of China
(No.62021002), Tsinghua BNRist and Beijing Key
Laboratory of Industrial Big Data System and Application,
Beijing Natural Science Foundation(QY25044).
\end{ack}

\bibliographystyle{plainnat}

\bibliography{references}

\appendix
\section{Backbone Models}
\label{app:backbones}

We evaluate \modelname on two open-source multimodal discrete diffusion language models, LLaDA-V~\citep{you2025lladavlargelanguagediffusion} and Lumina-DiMOO~\citep{xin2025luminadimooomnidiffusionlarge}. Although both use discrete diffusion as the decoding paradigm, they differ in how the visual modality enters the model, which is relevant for how the visual-ablation pass in \modelname is implemented.

\paragraph{LLaDA-V.}
LLaDA-V follows the visual instruction tuning paradigm and uses the LLaDA-8B-Instruct masked diffusion language model as its language tower. Visual inputs are first processed by a SigLIP\,2 vision tower (\textit{siglip2-so400m-patch14-384}, 384$\times$384 input, 729 visual tokens per image) and then projected into LLaDA's word embedding space by a randomly initialized two-layer MLP. The resulting visual tokens are concatenated as a visual prefix in front of the text tokens, and the language tower performs masked prediction over the answer tokens under bidirectional attention. Because visual inputs enter as continuous projected embeddings, the visual-ablation pass in \modelname is implemented by replacing those embeddings with the \texttt{[MASK]} token embedding while keeping the text tokens and the mask pattern unchanged.

\paragraph{Lumina-DiMOO.}
Lumina-DiMOO is a unified discrete diffusion model in which both text and images are represented as tokens drawn from a single joint vocabulary. It is initialized from LLaDA-Base without structural modifications and extended to multimodal inputs by adding 8,192 visual tokens from an aMUSEd-VQ image tokenizer (16$\times$16 downsampling) and a small set of special tokens (e.g., \texttt{<IMAGE>}, \texttt{</IMAGE>}, \texttt{<end-of-line>}). Each image is converted into a sequence of discrete visual tokens that are interleaved with text tokens in a single sequence, on which the model performs masked prediction under bidirectional full attention; for image understanding, decoding follows LLaDA's semi-autoregressive blockwise mask-predict schedule. Because the visual input is itself a token sequence rather than continuous embeddings, the visual-ablation pass in \modelname is implemented by replacing those visual tokens with \texttt{[MASK]} tokens while keeping the text tokens and the mask pattern unchanged.

\section{Decoding Procedure}
\label{app:decoding}

\subsection{\modelname Decoding Algorithm}
\label{app:algorithm_pseudocode}

Algorithm~\ref{alg:vigor} summarizes the full inference procedure. \modelname preserves the blockwise mask-predict schedule of the base model and only replaces the ranking rule used to select positions to unmask inside each block.

\begin{algorithm}[H]
\caption{\modelname decoding for a single block.}
\label{alg:vigor}
\begin{algorithmic}[1]
\Require prompt $q$, image $I$, current block $B$ of masked positions, per-step budget $k$, numerical stability constant $\varepsilon = 10^{-8}$
\While{$B$ contains masked positions}
  \State $p_{\mathrm{img}} \gets \mathrm{softmax}\bigl(f_\theta(q, I, \text{current mask})\bigr)$ \Comment{image-conditioned pass}
  \State $p_{\mathrm{noimg}} \gets \mathrm{softmax}\bigl(f_\theta(q, \emptyset, \text{current mask})\bigr)$ \Comment{visual-ablation pass}
  \For{each masked position $i \in B$}
    \State $\hat{y}_i \gets \arg\max_y p_{\mathrm{img}}(y \mid i)$
    \State $c_i^{\mathrm{img}} \gets p_{\mathrm{img}}(\hat{y}_i \mid i)$, \quad $c_i^{\mathrm{noimg}} \gets p_{\mathrm{noimg}}(\hat{y}_i \mid i)$
    \State $s_i \gets (c_i^{\mathrm{img}} - c_i^{\mathrm{noimg}}) + \bigl(\log(c_i^{\mathrm{img}} + \varepsilon) - \log(c_i^{\mathrm{noimg}} + \varepsilon)\bigr)$
  \EndFor
  \State $\mathcal{T} \gets \text{top-}k\bigl(\{s_i : i \in B\}\bigr)$
  \State Unmask each $i \in \mathcal{T}$ to $\hat{y}_i$; keep $B \setminus \mathcal{T}$ masked
\EndWhile
\end{algorithmic}
\end{algorithm}

The logarithmic term is computed as $\log p_{\mathrm{img}}(\hat{y}_i) - \log p_{\mathrm{noimg}}(\hat{y}_i)$ directly from log-softmax outputs, avoiding numerical underflow on low-probability tokens. The two forward passes share prompt tokens and the current mask pattern; only the image conditioning differs. Because \modelname does not modify model weights or the block schedule, the image-conditioned pass matches the vanilla baseline exactly, which isolates the effect of the reordering rule from any change in parameters or schedule.

\subsection{Model-Specific Implementation}
\label{app:implementation}

Applying \modelname to LLaDA-V and Lumina-DiMOO requires only a thin wrapper around each model's existing decoding loop. Table~\ref{tab:impl_details} summarizes the model-specific choices; all other hyperparameters (block size, per-step $k$, generation length, sampling temperature) are inherited from each model's default decoding configuration and match the vanilla baselines reported in Section~\ref{sec:experiments}.

\begin{table}[H]
\centering
\small
\caption{Model-specific implementation of \modelname on the two backbones.}
\label{tab:impl_details}
\begin{tabular}{lll}
\toprule
 & \textbf{LLaDA-V} & \textbf{Lumina-DiMOO} \\
\midrule
Image-token placement & visual prefix before text tokens & interleaved per chat template \\
Visual-ablation pass & image embeddings replaced by \texttt{[MASK]} & image tokens replaced by \texttt{[MASK]} \\
Mask token & official \texttt{[MASK]} id (unchanged) & official \texttt{[MASK]} id (unchanged) \\
Block schedule & identical to baseline & identical to baseline \\
Ranking rule & $s_i$ from Eq.~\ref{eq:vigor_score} & $s_i$ from Eq.~\ref{eq:vigor_score} \\
\bottomrule
\end{tabular}
\end{table}

\section{Adapting AR LVLM Inference-Time Baselines to Multimodal Discrete Diffusion Backbones}
\label{app:baselines}

Mask-diffusion decoding differs from autoregressive decoding in two ways that affect how the inference-time hallucination mitigation methods we use as baselines have to be adapted: every step predicts all masked positions in parallel rather than producing one new token, and the choice of which positions to commit at each step is itself driven by the visual condition through token confidence. This appendix describes how we adapt VCD~\citep{leng2024mitigating} and ICT~\citep{chen2025ict} to LLaDA-V and Lumina-DiMOO under these two properties.

\subsection{VCD}
\label{app:baselines:vcd}

Original VCD on AR LVLMs decodes one token per step. At each step it runs two forward passes, one with the clean image $I$ and one with a distorted image $I'$ (e.g., obtained from $I$ by Gaussian noise) on the \emph{same} prefix, and contrasts the two logit vectors at the single decoding position:
\begin{equation}
s_v \;=\; (1+\alpha)\, f_\theta(v \mid I, q, \mathbf{z}^{(s)}) \;-\; \alpha\, f_\theta(v \mid I', q, \mathbf{z}^{(s)}),
\label{eq:vcd}
\end{equation}
where $v \in \mathcal{V}$ is a candidate token, $\mathbf{z}^{(s)}$ is the prefix decoded so far at AR step $s$ (in the notation of Section~\ref{sec:method}), $f_\theta(\cdot)$ denotes pre-softmax logits as in Algorithm~\ref{alg:vigor}, and $\alpha > 0$ is the VCD contrast strength. An adaptive plausibility cutoff additionally keeps $v$ only if $f_\theta(v \mid I, q, \mathbf{z}^{(s)}) \ge \log\beta + \max_{v' \in \mathcal{V}} f_\theta(v' \mid I, q, \mathbf{z}^{(s)})$, where $\beta \in (0,1]$ is a cutoff threshold. Two properties of the AR setting are essential: there is exactly one position to be decoded at this step, and the two passes share the prefix automatically.

In each backbone's blockwise mask-diffusion loop, our adaptation follows the same blockwise mask-predict schedule as the vanilla baseline and applies the contrast in Eq.~\ref{eq:vcd} at every denoising step. Three implementation choices are needed to make the per-position contrast well-defined under parallel mask-predict: the two branches are initialized to the same masked sequence and prefix tokens, the same set of transferred tokens is copied into both branches at every denoising step, and the commit decision is driven by the clean side alone. The dual visual condition enters where each backbone naturally consumes vision: at the projected embedding level for LLaDA-V, and at the visual-token level for Lumina-DiMOO.

\subsection{ICT}
\label{app:baselines:ict}

Original ICT trains per-head binary probes on a calibration set of (clean image, blurred image) pairs to identify attention heads that linearly separate visually-grounded from prior-driven activations, selects the top-$K$ such heads, and at each AR decoding step adds the per-head class-mean activation shift $S_\ell^n$ (the mean difference between trusted and untrusted activations on the calibration set), scaled by an intervention strength $\gamma > 0$, to the per-head input of \texttt{self\_attn.o\_proj}, but only at the position of the right-most newly-generated token:
\begin{equation}
h_\ell^n \;\leftarrow\; h_\ell^n \;+\; \gamma \cdot S_\ell^n,
\qquad (\ell, n) \in \text{top-}K,
\label{eq:ict}
\end{equation}
where $\ell \in \{1,\ldots,L\}$ and $n \in \{1,\ldots,N\}$ index the $L$ transformer layers and the $N$ attention heads per layer respectively, $h_\ell^n$ is the per-head input to \texttt{self\_attn.o\_proj} at $(\ell, n)$, and $K \le LN$ is the number of selected heads.

Since a dLLM has no single ``right-most newly-generated token'' at any step (the generation span is a mixture of already-committed and still-masked positions, all updated in parallel), our adaptation widens the intervention scope to the entire span of generation positions. We run ICT in its default configuration with both image-level and object-level interventions enabled, with the per-head activation shift vectors $S_\ell^n$ in Eq.~\ref{eq:ict} estimated separately for each backbone on the same calibration data as the original paper.

\section{Inference Cost}
\label{app:inference_cost}

\begin{table}[H]
\centering
\small
\caption{Decoding time (ms) on A100 GPUs at generation lengths 32/64/128. The \emph{Ratio} column shows \modelname vs.\ its baseline; the ratio cell is highlighted in \caplite{green}.}
\label{tab:time_overhead}
\renewcommand{\arraystretch}{1.05}
\setlength{\tabcolsep}{6pt}
\begin{tabular*}{\linewidth}{@{\extracolsep{\fill}}lcccc@{}}
\toprule
\textbf{Method} & \textbf{len 32} & \textbf{len 64} & \textbf{len 128} & \textbf{Ratio} \\
\midrule
LLaDA-V                                    & 3644.6  & 7338.2  & 15552.8 & --           \\
\hspace{0.8em}+\,\modelname                & 7417.3  & 14954.1 & 31684.9 & \bestnum{$\times 2.0$} \\
\addlinespace[1pt]
Lumina-DiMOO                               & 6395.4  & 13430.5 & 27136.1 & --           \\
\hspace{0.8em}+\,\modelname                & 13129.4 & 27505.8 & 55535.7 & \bestnum{$\times 2.0$} \\
\bottomrule
\end{tabular*}
\end{table}

\begin{table}[H]
\centering
\small
\caption{Memory usage (GiB) on Lumina-DiMOO. Delta is reported as \modelname minus baseline and computed from unrounded measurements.}
\label{tab:memory_overhead}
\setlength{\tabcolsep}{8pt}
\begin{tabular}{lccc}
\toprule
\textbf{Metric} & \textbf{Baseline} & \textbf{\modelname} & \textbf{Delta} \\
\midrule
Peak alloc. & 29.56 & 34.93 & $+5.36$ \\
Peak reserved & 78.74 & 78.61 & $-0.13$ \\
\bottomrule
\end{tabular}
\end{table}

Table~\ref{tab:time_overhead} shows that \modelname roughly doubles decoding time on both LLaDA-V and Lumina-DiMOO across all tested lengths, since each denoising step adds a visual-ablation forward pass. Table~\ref{tab:memory_overhead} further shows that, on Lumina-DiMOO, \modelname increases peak allocated memory by $5.36$ GiB while leaving peak reserved memory essentially unchanged. The $\times 2.0$ time factor is stable across both model scales and all tested generation lengths, indicating a clean multiplicative cost. This is the price for a training-free, plug-and-play reordering rule that requires no retraining, extra supervision, or architectural change.

\section{Appendix Case Study}
\label{app:case_study}

\begin{figure}[!htbp]
\centering
\includegraphics[width=\linewidth,page=1]{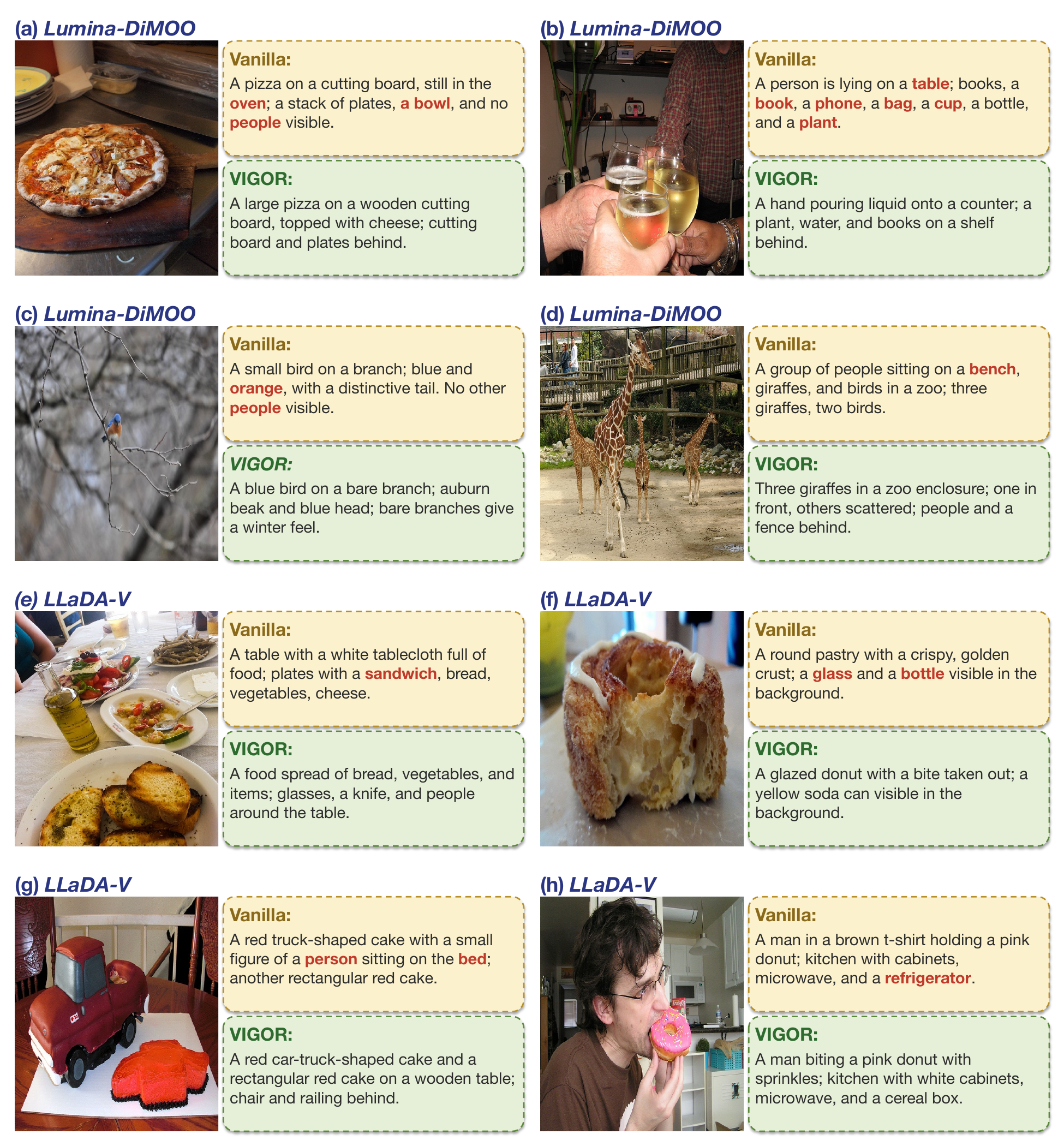}
\caption{Appendix case study. Eight image--caption comparisons between baseline and \modelname decoding on Lumina-DiMOO (a--d) and LLaDA-V (f--h). Hallucinated object words are highlighted in \textcolor{red}{\textbf{red}}.}
\label{fig:appendix_case_study}
\end{figure}

Figure~\ref{fig:appendix_case_study} shows eight case studies for Lumina-DiMOO (a--e) and LLaDA-V (f--h). For each image, the baseline caption (top, cream) and the \modelname caption (bottom, sage) are produced by the \emph{same} model under the \emph{same} generation length and block schedule; \modelname only reorders the commitment of masked positions at inference. Hallucinated object words in the baseline caption are highlighted in \textcolor{red}{\textbf{red}}. On Lumina-DiMOO, the baseline invents an \emph{oven} and \emph{bowl} around a pizza (a), fabricates a tabletop scattered with \emph{books, phone, bag, cup, and plant} (b), turns a winter bird scene into something with an \emph{orange} bird and \emph{people} (c), inserts a \emph{bench} into a zoo scene (d), and adds a \emph{sandwich} to a banquet table (e); in every case \modelname removes the invented objects while still describing the visible scene. On LLaDA-V, the baseline imagines a \emph{glass} and \emph{bottle} behind a donut that is actually next to a soda can (f), hallucinates a \emph{person} sitting on a \emph{bed} on a truck-shaped cake (g), and adds a \emph{refrigerator} to a kitchen that has only cabinets and a microwave (h); \modelname keeps to the visible objects in each case. \modelname captions are not shorter than the baselines, indicating that the improvement does not come from producing less detailed descriptions.

\end{document}